\documentclass{article}
\usepackage{todonotes}
\usepackage{pgfplots}
\usepackage{url}
\usepackage{myexample}
\usepackage{newunicodechar}
\usepackage{natbib}
\newunicodechar{€}{euro}

\title{A corpus of precise natural textual entailment problems}
\author{Jean-Philippe Bernardy \and Stergios Chatzikyriakidis}
\newcommand\Example[3]{%
  \textbf{Example:} (Problem #1) \\
  \textit{
  P: #2 \\
  H: #3
}}

\begin{document}

\maketitle{}

\begin{abstract}
  In this paper, we present a new corpus of entailment problems. This corpus combines the following characteristics: 1. it is precise (does not leave out implicit hypotheses) 2. it is based on ``real-world'' texts (i.e. most of the premises were written for purposes other than testing textual entailment). 3. its size is 150.

  The corpus was constructed by taking problems from the Real Text Entailment and discovering missing hypotheses using a crowd of experts. We believe that this corpus constitutes a first step towards wide-coverage testing of precise natural-language inference systems.
\end{abstract}

\section{Intro}
Reasoning is part of our every day routine:  we hear Natural  Language (NL) sentences,  we participate in dialogues, we  read books or legal documents. Successfully understanding, participating or communicating with others in these situations presupposes some form of reasoning: about individual sentences, whole paragraphs of legal documents, small or bigger pieces of dialogue and so on. The human reasoning performed in these different situations cannot be explained by a single rigid system of reasoning, plainly because reasoning is performed in different ways in each one of them.  Consider the following example:

\myexample{Three representatives are needed.}

If a human reasoner with expert knowledge was to interpret the above utterance in a legal context, s/he will most probably judge that a situation where more than three references are provided could be compatible with the semantics of the utterance. To the contrary, if the same reasoner was to interpret the above  as part of a casual, everyday conversation, then \textit{three} would most likely be interpreted as \textit{exactly three}, making the same situation incompatible with the utterance. In this paper, we want to focus on precise inference, inference which is  either performed by experts or normal people after taking some time to consider the inferences that follow or not from a set of premises. The problem one encounters in testing   systems of this type of reasoning is the fact that no large scale datasets of this sort of inference exist. The commonly used datasets for systems fit for this type of precise reasoning, i.e. logical systems based on some model of formal semantics for Natural Language (NL), are the FraCaS test suite and the Recognizing Textual Entailment (RTE) datasets.\footnote{To a certain extent, the SICK dataset~\cite{sick} can be also thought to be fit for testing logical approaches. However, given that SICK has been originally designed to test distributional compositional semantics approaches, it is not an optimal choice for our task.} 

\subsection{The FraCaS test suite}
The  FraCaS test suite\footnote{\url{ftp://ftp.cogsci.ed.ac.uk/pub/FRACAS/del16.ps.gz}} is an NLI data set consisting of 346 inference problems. Each problem contains 
one or more premises followed by one yes/no-question.\footnote{Yet FraCaS exhibits some formal problems. For example, four problems are not formulated as a question.} There is a three way classification: YES, NO or UNK (unknown, see Figure 1 for an example from FraCaS). The FraCaS test suite was later on turned into machine-readable format by Bill McCartney\footnote{\url{www-nlp.stanford.edu/~wcmac/downloads/fracas.xml}. There, the conclusion is presented in two forms: one in the original question format and one as a declarative statement following from the premises (see (\ref{Fracas})).}

Expansions of FraCaS include: a) MultiFraCaS, in effect a multilingual FraCaS\footnote{\url{www.ling.gu.se/~cooper/multifracas/}}, and b) JSem, the Japanese counterpart to  FraCaS, which expands the original FraCaS  in a number of ways.\footnote{More info on the suite and its innovations compared to the original FraCaS can be found here: \url{http://researchmap.jp/community-inf/JSeM/?lang=english}.} 

Even though the FraCaS test suite contains a rather small number of  examples (346), it  covers a lot of NLI cases and is, at least to some extent, multilingual. It is to some extent precise, even though there are test cases that do not involve a clear answer and thus are dubbed as undefined in Bill MacCartney's XML version. A further drawback of the FraCaS test stuite is that it involves constructed examples, rather than real text.

\myexample{An UNK example from the FraCaS test suite.
	\label{Fracas}
	\begin{itemize} \item[\textbf{P1}] A Scandinavian won the Nobel Prize.
		
		\item[\textbf{P2}]	Every Swede is  Scandinavian.
		
		\item[\textbf{H.}] Did a Swede win the Nobel prize? 
		
		\item[\textbf{H.}] A Swede won the Nobel prize.
		
		\item[\textbf{Label}] 	UNK [FraCaS 065] 	\end{itemize}
	
}

\subsection{Recognizing Textual Entailment}

The Recognizing Textual Entailment (RTE) challenges first appeared in 2004 as a means to test textual entailment, i.e. relations between a premise text and a hypothesis text (\ref{RTE}):

\myexample{ An entailment example from  RTE1. \label{RTE}
	\begin{itemize}
		\item[\textbf{P.}]	Budapest again became the focus of
		national political drama in the late 1980s,
		when Hungary led the reform movement
		in eastern Europe that broke the
		communist monopoly on political power
		and ushered in the possibility of multiparty
		politics.
		
		\item[\textbf{H.}] In the late 1980s Budapest became the
		center of the reform movement. 
		
		\item[\textbf{Label}] 	Entailment [RTE702] 	\end{itemize}

}In contrast to the FraCaS test suite, the RTE challenges use naturally occurring data as premises. The hypothesis text is then constructed based on this premise text.  There is either a binary or a tripartite classification of entailment --- depending on the version of RTE.  The first two RTE challenges follow the former scheme and make a binary classification of entailment (entailed or not entailed).   Tripartite classification  (entailment,  negation of the hypothesis entailment or no entailment) is added in the later datasets, retaining  two way classification versions as well. Seven RTE challenges have been created altogether. 

The main advantages of the RTE challenges is their use of examples from natural text and the inclusion of cases  that require presupposed information, mostly world knowledge. Indeed, the very \emph{definition} of inference assumed in a number of the examples is problematic. As \citet{ZaenenKarttunenCrouch2005} have pointed out,
RTE platforms suffer from cases of inference that should not be categorized as such. For these cases, a
vast amount of world knowledge needs to be taken into consideration (that most importantly not every
linguistic agent has). In this paper, and having the RTE as our starting point  challenges, we claim that RTE is insufficiently precise to perform logical reasoning or precise reasoning tasks and we take up the task of validating our working hypothesis and proposing a method for doing proper collection of precise entailment pairs in the style of RTE. Of course, the creators of RTE had in mind a more loose definition of inference where both a precise and an imprecise definition of entailment would be at play. \citet{dagan:2010} mention that ``our applied notion of textual entailment is also related, of course, to classical
semantic entailment in the linguistics literature... a common definition of entailment
 specifies that a text t entails another
text h (hypothesis, in our terminology) if h is true in every circumstance (possible
world) in which t is true." This is close to what we want to capture in this paper. But, at the same time, \citet{dagan:2010} also mention that ``however, our applied definition allows for cases in which the truth of the hypothesis
is highly plausible, for most practical purposes, rather than certain". It is these   cases we want to make more precise, in the sense of making the supporting hidden inferences that are at play in many of the RTE examples explicit. In a way, what we are aiming at is a methodology of constructing  entailment datasets in the style of RTE, that will involve a more precise definition of entailment and will further record any missing/hidden premises are used in justifying or not an entailment pattern.





\section{Method}

We have randomly selected 150 problems out of the RTE corpus which
were marked as ``YES'' (i.e. entailment holds). The problems were not
further selected nor doctored by us.  The problems were then re-rated
by experts in logic and/or linguistics. For each problem, three
experts were consulted, and each expert rated 30 problems. More
precisely, the experts were instructed to re-consider each problem and
be especially wary of missing hypotheses. If they considered the
entailment to hold, we still gave the instruction to optionally
mention any additional implicit hypothesis that they would be using.
Similarly, if they considered that there was no entailment in the
problem, they were given prompted to (optionally) give an argument for
their judgement.

In order to facilitate data collection, the experts were chosen from
the network of contacts of the author. Despite this method, the
process of data collection took nearly six months. The authors
themselves were put to contribution in the data-collection process
(taking one set of 30 problems each) in order to complete the survey.

Additionally, collecting all the inputs received and using our best
judgments, we have put together a test set of 150 problems comprised
of the original problems, a new judgement (``yes'' or ``no''), and
added missing hypotheses (if ``yes'' is a reasonable option).

\section{Results}

In the process, we have gather a total of 449 expert judgments (one
expert failed to answer a given problem), 146 missing hypotheses and
47 explanations for negative judgments.

\begin{table}
  \centering
  \begin{tabular}{lrr}
    Type & Number & Ratio over total \\
    \hline
    Yes, with no missing hypothesis & 223 & 0.49 \\
    Yes, with missing hypotheses & 146 & 0.33 \\
    No, with no explanation & 33 & 0.07 \\
    No, with explanation & 47 & 0.10 \\
    \hline
    Total of doubtful entailment & 226 & 0.50 \\
  \end{tabular}
  \caption{Number of responses by type}
  \label{tab:number-by-type}
\end{table}
Despite being marked as ``yes'', the problems with a reported missing
hypothesis should really be classified as ``no'', if one does not
assume external knowledge. (See below for further discussion on the
reported missing hypotheses.)  Thus crunching the numbers, we see that
more that half of the responses express some doubt about
entailment. Remember that all problems were marked as ``yes'' by the
creators of the RTE3 testsuite --- we find here that one average, one
expert in two is likely to cast a doubt over this ``yes''.

However, each problem was classified by three experts. The histogram
below shows the distribution of number of experts casting doubt on
entailment, over all problems.

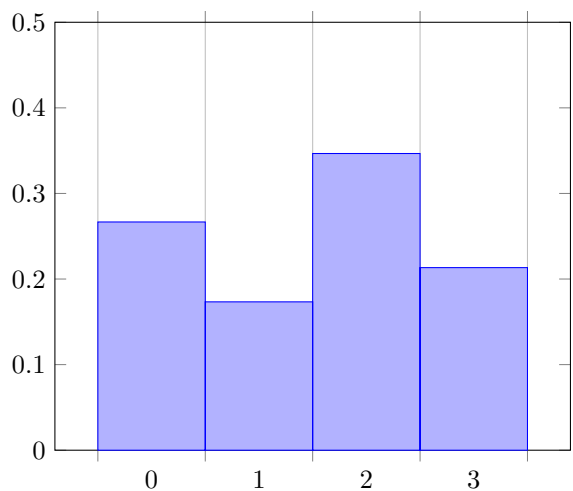
\begin{center}
\begin{tikzpicture}
\begin{axis}[ybar interval, ymax=0.5,ymin=0, minor y tick num = 0]
\addplot coordinates { (0, 0.26666666666666666) (1, 0.17333333333333334) (2, 0.3466666666666667) (3, 0.21333333333333335) (4,0) };
\end{axis}
\end{tikzpicture}
\end{center}
Unfortunately we can only draw preliminary conclusions, due to the
limited number of respondents for each problem. However, we can make the following observations:
\begin{enumerate}
\item Perfect agreement (0 or 3 doubts) occur in 48 percent of cases.
\item The probability of having a single doubt being cast is the lowest.
\end{enumerate}
We find this level of agreement indicative of a good level of
reliability. Additionally, with three experts per problem, we are very
likely to discover most missing hypotheses and incorrect entailments.

In our compilation of answers, we have marked 42 problems as straight
``No'', 64 as ``Yes'' with missing implicit hypotheses and ``44'' as
plain ``Yes''. This means that, we expect, in our opinion, 28\% of
problems to be incorrectly labeled in RTE3 \emph{even assuming
  reasonable world knowledge.} An additional 42\% of problems require
additional (yet reasonable to assume) hypotheses for entailment to
hold formally, as prescribed by RTE3. This leaves only 30\% of
problems to acceptable as such. The reason that the amount of doubt is
larger than in the average numbers quoted above is that, for many
problems, certain missing hypotheses and/or error were not detected by
a majority experts, but, after careful inspection, we judge that the
minority report is justified.

We have additionally tagged each missing hypothesis according to
the following classification:
\begin{enumerate}
\item Linguistic subtleties (Labeled ``Language''; Example:
  ``ownership in the past is enough to justify the possessive in the
  present''; 9 occurrences in our sample)
\item Lexical meaning, sometimes specific to the context of the
  problem; (Labeled ``Lexicon''; Example: ``buying entails selling''; 15 occurrences in our sample)
\item World knowledge (Labeled ``World''; Example: ``Increased amounts of CO2 and other
  greenhouse gases cause Greenhouse effect.''; 13 occurrences in our sample)
\item Other missing hypothesis, see below for further details.
\end{enumerate}

\subsection{``Yes if ...'' vs ``No because ...''?}

The classification between ``yes'' with missing hypotheses and ``no''
is sometimes a tenuous one --- which is why we elected to group those
categories in our summaries above. Indeed, consider the following example:

\Example{672}{%
P: Philip Morris the US food and tobacco group
that  makes  Marlboro,   the  world’s  best-selling
cigarette, shrugged off strong anti-smoking senti-
ment in the US.}{%
H: Philip Morris owns the Marlboro brand.
}

We got the following answers:
\begin{description}
\item[A1] Yes, if making involves owning the brand
\item[A2] Yes, if making something implies owning the brand
\item[A3] No, because making the product does not imply owning the brand
\end{description}
It is clear for all experts that a premise is missing, but some will
consider it acceptable to add, others will not.

\subsection{Analysis of reported missing hypotheses and incorrect labeling in RTE3}

While most errors in the original RTE classification can not be attributed

\paragraph{Pragmatic Strengthening}
It appears that wrong conclusions are sometimes justified by appeal to
pragmatic strengthening of the premises. (Many problems, at least 175, 50, 51, 454, 643,
722, 740, 278 in our sample).  Indeed, in our experts judgments, we have
found cases where problems were marked as ``yes'' in RTE, and seem to have been implicitly justified with
a hypothesis which is, taken in isolation, false, but which could make
sense in the context of the premises.

This can be problematic in the context of an entailment
system. Indeed, the problem does not become ``is there entailment'',
but rather ``does the questioner intend entailment''. To give an example, take a look at the following: 

\Example{454}{%
On Aug. 6, 1945, an atomic bomb was exploded on Hiroshima with an estimated equivalent explosive force of 12,500 tons of TNT, followed three days later by a second, more powerful, bomb on Nagasaki.
}{%
In 1945, an atomic bomb was dropped on Hiroshima.
}
(Bombs can explode without being dropped.)

The above example was marked as an entailment in the original RTE suite. However, one of our annotators marked as a non-entailment, providing the justification one sees in parentheses. The justification is of course correct. However, one can also claim that given the context of the text, and the fact that one can take dropping of the atomic bombs in Hiroshima and Nagashaki as a recoverable world-knoweldge premise, Yes is also an option. But not an option, if precise inference systems are to be trained. 

\paragraph{Mistaking claims for truth}
The single one largest single specific source of incorrect labeling,
found in 12 problems in our sample (750, 754, 756, 757, 51, 66, 178,
225, 294, 588,643,659) out of 42 errors, is mistaking claims for
truth, as in the following example.

\Example{294}{%
Mental health problems in children and adolescents are on the rise,
the British Medical Association has warned, and services are
ill-equipped to cope.}{%
Mental health problems increase in the young.
}

As it should be obvious with a instant's thought, the above should
entail only if the word of the British Medical Association can be
taken for fact. While it may be safe to behave as such in many
situations in the real world, one cannot do so when reasoning
precisely.

\paragraph{Mistaking intentions and facts}
Another source of common mistakes is the confusion of intentions and
facts, found in 8 problems in our sample (33,148,191,396,420,59,121,166).

\Example{191}{%
Though Wilkins and his family settled quickly in Italy, it wasn't a
successful era for Milan, and Wilkins was allowed to leave in 1987 to
join French outfit Paris Saint-Germain.}{%
Wilkins departed Milan in 1987.
}
(Even though Wilkins was allowed to leave, it does not mean he actually left.)

\paragraph{Mistaking the past for the present}

Rather obviously, events described in the past cannot be taken to hold
currently.  Yet this error is still found in 6 problems in our sample
(255,230,118,308,454,175).

\Example{308}{%
On 29 June the Dutch right-wing coalition government collapsed. It
was made up of the Christian-democrats (CDA) led by Prime Minister Jan
Peter Balkenende, the right wing liberal party (VVD) and the so-called
'left-liberal' D66.
}{%
Three parties form a Dutch coalition government.
}

(The coalition may have collapsed at the time of solving the problem)

\paragraph{Incorrect  reasoning with lexical semantics}
The final common source of errors that we identify is incorrect
reasoning with lexical semantics.  This error a bit more subtle than the
others, but found only five
occurrences in our sample (59,202,221,231,463). One explanation for its
relatively low frequency is that RTE subject paid special attention to
it. To give an example, consider the following:

\Example{463}{%
 Catastrophic floods in Europe endanger lives and cause human tragedy as well as heavy economic losses
}{%
Flooding in Europe causes major economic losses.}

In the above example, incorrect reasoning with adjectival semantics is made. Catastrophic floods forms a subset of all floods. Thus, the hypothesis does not follow, basically because not all floods are catastrophic. 

\subsection{Use of world-knowledge}

We find that entailment problems which depend on any non-trivial
amount of world knowledge are problematic from the point of view of
training and testing systems for entailment.  Indeed, in the presence
of a large number of arbitrary facts, the conclusion can come solely
from such knowledge, completely ignoring the premise. At best, the
premise is serving as priming the memory of the reader. We find this
issue to happen in RTE in a significant number of cases, including
some of those listed above. For example, in problem 454, it is common
knowledge that a bomb was indeed dropped on Hiroshima --- yet we may
consider that the entailment does not hold \emph{as such}.

\section{Conclusion and Future work}

We find that our hypothesis is validated: RTE is not suitable as such
to test a precise NLI system, because, for entailment to hold as
tagged in RTE, much world-knowledge is required and many missing
hypotheses are omitted.

By using a crowd of experts to repair the missing hypotheses, we have
constructed a dataset of 150 precise entailment problems, based on
text found in real-world corpora. Even though the dataset is on the
small size, it is, to the best of our knowledge, the first of this
kind.

An issue with the method that we used to construct our new dataset of
entailment problems is that it is difficult to scale: it demands
several minutes of scarce expert work per constructed problem.  Our
plan is to investigate the possibility to gamify the process, so that
lots of people can participate in the construction of precise
entailment problems, as a form of entertainment. We leave any detail
to further work, but this would be an asymmetric game, where one one
player tries construct watertight entailment problems, and the
opposing player would try and refute such entailment problems (say, by
giving counter-examples). Identifying the possible kind of mistakes,
as we have done here, will help prompting the players about things to
look for --- in either of the possible roles.

Finally we find it striking that many mistakes committed in RTE are
falling for classical fallacies (appealing to authority accounts for
more than a quarter of errors). 
Thus, we believe that while the entailment problems are useful for the
construction of NLI systems, the added hypotheses (or moves taken in
our hypothetical game) could be of interest to the linguistic
community at large.


\section*{Acknowledgments}

We are grateful to the crowd of experts that performed the hard work
of precisely annotating problems. Most of them chose to remain
anonymous. The others were, in alphabetical order: Rasmus Blank, Robin
Cooper, Matthew Gotham, Julian Hough and Aarne Talman.

\bibliographystyle{plainnat}
\bibliography{NLI}

\section*{Test suite}
We join the complete compiled test suite below. Please note that the
problem numbers are not consecutive, because we chose them to be their
identifier in the RTE3 test suite.
\vspace{1cm}

{\small

\textbf{Problem 4}\\
P: "The Extra Girl" (1923) is a story of a small-town girl, Sue Graham (played by Mabel Normand) who comes to Hollywood to be in the pictures. This Mabel Normand vehicle, produced by Mack Sennett, followed earlier films about the film industry and also paved the way for later films about Hollywood, such as King Vidor's "Show People" (1928)\\
H: "The Extra Girl" was produced by Sennett\\
A: Yes\\
Lexicon: In this context, 'vehicle' is a paraphrase for story/film\\
Nitpick: Sennett is taken as shorthand for Mack Sennett\\

\textbf{Problem 17}\\
P: Allen was renowned for his skill at scratch-building and creating scenery, and he pioneered the technique of weathering his models to make them look old and more realistic\\
H: Allen introduced a new technique of creating realistic scenery\\
A: Yes\\

\textbf{Problem 24}\\
P: Bountiful arrived after war's end, sailing into San Francisco Bay 21 August 1945. Bountiful was then assigned as hospital ship at Yokosuka, Japan, departing San Francisco 1 November 1945\\
H: Bountiful reached San Francisco in August 1945\\
A: Yes\\

\textbf{Problem 26}\\
P: The Prime Minister of Spain Zapatero visited Brazil, Argentina, Chile and Uruguay recently, in a effort to build a left axis in South America. The cited countries' South American Presidents agreed to collaborate at international level, particularly in the United Nations , European Union and with Paris, Berlin and Madrid\\
H: Brazil is part of the United Nations\\
A: Yes\\
Missing: Collaborating in the UN requires being part of the UN\\

\textbf{Problem 27}\\
P: Under the headline "Greed instead of quality", Germany's Die Tageszeitung says no good will come of the acquisition of the publisher Berliner Verlag by two British and US-based investment funds\\
H: British and US-based investment funds acquire Berliner Verlag\\
A: No; because there is no evidence the planned takeover in the premises was actually carried out\\

\textbf{Problem 32}\\
P: Carl Smith collided with a concrete lamp-post while skating and suffered a skull fracture that caused a coma. When he failed to regain consciousness, his parents on August 8 consented to his life support machine being turned off\\
H: Carl Smith died on August 8\\
A: Yes\\
Missing: Turning off a life support machine always results in death of the connected patient\\
Missing: The life support machine was turned off on the same day as the parents gave the consent\\

\textbf{Problem 33}\\
P: As leaders gather in Argentina ahead of this weekends regional talks, Hugo Chávez, Venezuela's populist president, is using an energy windfall to win friends and promote his vision of 21st-century socialism\\
H: Chávez is a follower of socialism\\
A: Yes\\
Missing: His vision of 21st-century socialism can be classified as socialism\\
Missing: Promoting a vision of X makes one a follower of X\\

\textbf{Problem 44}\\
P: A former employee of the company, David Vance of South Portland, said Hooper spent a lot of time on the road, often meeting with customers between Portland and Kittery\\
H: David Vance lives in South Portland\\
A: Yes\\

\textbf{Problem 49}\\
P: The British government did not initially purchase the weapon and civilian sales were modest. However the U.S. Civil War began in 1860 and the governments of both the United States and the Confederacy began purchasing arms in Britain\\
H: During the Civil War the government of the United States bought arms from Britain\\
A: Yes\\
Missing: They have not have been purchasing arms from someone else in Britain\\

\textbf{Problem 50}\\
P: Edison decided to call "his" invention the Kinetoscope, combining the Greek root words "kineto" (movement), and "scopos" ("to view")\\
H: Edison invented the Kinetoscope\\
A: No\\
Reason: The quotes around “his” seem to indicate that it is not sure\\

\textbf{Problem 51}\\
P: "I want to go back again. But I am afraid, honestly, I am afraid. Propaganda against me made people think I am terrorist.", said el-Nashar\\
H: El-Nashar is accused of terrorism\\
A: No\\
Reason: Propaganda is not necessarily about terrorism\\
Claim: El-Nashar is truthful\\

\textbf{Problem 57}\\
P: FermiLab's goal is to ensure that if a program runs and is certified on RedHat Enterprise, then it will run on the corresponding Fermi Linux LTS release. They have built Fermi Linux LTS for Fermilab's use, and that is their ultimate goal\\
H: FermiLab created Linux LTS\\
A: Yes\\
Lexicon: “Fermi Linux LTS” can be referred using the term “Linux LTS”\\

\textbf{Problem 58}\\
P: On the morning of 1 June, there was a blackout throughout most of the capital caused by urban commandos of the Farabundo Marti National Liberation Front (FMLN)\\
H: FMLN caused a blackout in the capital\\
A: Yes\\
Missing: The urban commandos were actually acting on behalf of FMLN\\

\textbf{Problem 59}\\
P: After graduating in 1977, Gallager chose to accept a full scholarship to play football for Temple University\\
H: Gallager attended Temple University\\
A: No\\
Reason: he might possibly have graduated from somewhere else, or might have decided to play for someone else even after initially accepting the scholarship\\

\textbf{Problem 66}\\
P: The Health Department has been made aware that property located at Bear Creek is under investigation by the NC Division of Waste Management for the illegal discharge of hazardous chemical materials\\
H: Hazardous chemical materials have been discovered at Bear Creek\\
A: Yes\\
Missing: The illegal discharge has been confirmed\\

\textbf{Problem 71}\\
P: As leaders gather in Argentina ahead of this weekends regional talks, Hugo Chávez, Venezuela's populist president is using an energy windfall to win friends and promote his vision of 21st-century socialism\\
H: Hugo Chávez acts as Venezuela's president\\
A: Yes\\
Missing: Being president entails acting as president\\

\textbf{Problem 73}\\
P: On October 1 2001, EU and other countries introduced the option for domestic animal owners to apply for Pet passports under the Pets Travel Scheme (PETS for short), for pets returning from abroad to the United Kingdom. This replaced the old system of 6 months compulsory quarantine for all domestic pets\\
H: In 2001, the EU introduced a passport for pets\\
A: No\\
Reason: the passport could have existed before, used for other reasons\\

\textbf{Problem 82}\\
P: Jerry Reinsdorf (born February 25 1936 in Brooklyn, New York) is the owner of Chicago White Sox and the Chicago Bulls. Recently, he helped the White Sox win the 2005 World Series and, in the process, collected his seventh championship ring overall (the first six were all with the Bulls in the 1990s), becoming the third owner in the history of North American sports to win a championship in two different sports\\
H: Jerry Reinsdorf has won 7 championships\\
A: Yes\\
Missing: Collecting a championship ring implies winning the championship\\

\textbf{Problem 92}\\
P: The Kinston Indians are a minor league baseball team in Kinston, North Carolina. The team, a Class A affiliate of the Cleveland Indians, plays in the Carolina League\\
H: Kinston Indians participate in the Carolina League\\
A: Yes\\

\textbf{Problem 99}\\
P: The Extra Girl (1923) is a story of a small-town girl, Sue Graham (played by Mabel Normand) who comes to Hollywood to be in the pictures. This Mabel Normand vehicle, produced by Mack Sennett, followed earlier films about the film industry and also paved the way for later films about Hollywood, such as King Vidors Show People (1928)\\
H: Mabel Normand starred in The Extra Girl\\
A: Yes\\

\textbf{Problem 107}\\
P: Since joining the Key to the Cure campaign three years ago, Mercedes-Benz has donated over \$2 million toward finding new detection methods, treatments and cures for women's cancers\\
H: Mercedez-Benz supports the Key to the Cure campaign\\
A: Yes\\
Missing: Donating money is supporting\\

\textbf{Problem 109}\\
P: ASCAP is a membership association of more than 200,000 U.S. composers, songwriters, lyricists and music publishers of every kind of music\\
H: More than 200,000 U.S. composers, songwriters, lyricists and music publishers are members of ASCAP\\
A: Yes\\

\textbf{Problem 115}\\
P: A light blue 1975 Ford Escort GL once owned by Pope John Paul II sold for \$690,000 Saturday to a Houston multimillionaire who said he plans to put it in a museum he wants to build in his hometown\\
H: A Houston multimillionaire buys the Pope's Ford Escort\\
A: Yes\\
Missing: John Paul II only owns a single Ford Escort\\
Language: ownership in the past is enough to justify the possessive in the present\\

\textbf{Problem 118}\\
P: According to Nelson Beavers, who is a co-owner of the current company, Carolina Analytical Laboratories, LLC. and has ownership/employment history with Woodson-Tenent and Eurofins, the septic system was installed in the early 1990s\\
H: Nelson Beavers is one of the owners of Carolina Analytical Laboratories\\
A: No\\
Reason: The mention of time indicates that the ownership may not be still current\\

\textbf{Problem 121}\\
P: Côte d'Ivoire's President Laurent Gbagbo promulgated new election laws on July 14, including the creation of an independent electoral commission to oversee the presidential vote, which is slated for October 30\\
H: New elections in Côte d'Ivoire will take place on October 30\\
A: No\\
Reason: What is slated may not come to pass\\

\textbf{Problem 122}\\
P: Nival was founded in 1996 by Sergey Orlovskiy. In early 2005, the company was bought by Ener1 Group, a Florida-based holdings company, for around US\$10 million\\
H: Nival was sold in 2005\\
A: Yes\\
Lexicon: Nival is a company\\
Lexicon: buying entails selling\\

\textbf{Problem 127}\\
P: As late as 1799, priests were still being imprisoned or deported to penal colonies and persecution only worsened after the French army led by General Louis Alexandre Berthier captured Rome and imprisoned Pope Pius VI, who would die in captivity in Valence, Drôme, France in August of 1799\\
H: Pope Pius VI died in France\\
A: Yes\\

\textbf{Problem 129}\\
P: Take consumer products giant Procter and Gamble. Even with a \$1.8 billion Research and Development budget, it still manages 500 active partnerships each year, many of them with small companies\\
H: Procter and Gamble spends \$1.8 billion for Research and Development\\
A: Yes\\
Missing: Its entire R\&D budget is spent\\

\textbf{Problem 132}\\
P: The president Cristiani spoke today at the El Salvador military airport before he left for Costa Rica to attend the inauguration ceremony of president-elect Rafael Calderon Fournier\\
H: Rafael Calderon Fournier has been elected president of Costa Rica\\
A: Yes\\
Lexicon: The phrase “president-elect” always indicates a president who has won a recent election\\
World: A presidential inauguration ceremony always takes place in the country of which the president was elected\\

\textbf{Problem 133}\\
P: From 1016 to 1030 the Normans were pure mercenaries, serving either Byzantines or Lombards, and then Sergius of Naples, by installing their leader Rainulf in the fortress of Aversa in 1030, gave them their first pied-à-terre and they began an organized conquest of the land\\
H: Rainulf was the leader of the Normans\\
A: Yes\\
Language: “their” takes “the Normans” as its antecedent\\

\textbf{Problem 142}\\
P: They named themselves for Saint Joan of Arc. The brigade began with 17 women, but soon grew to 135 members. Its mission was to obtain money, weapons, provisions, and information for the combatant men. Many smuggled weapons into the combat zones by carrying them in carts filled with grain or cement\\
H: Saint Joan of Arc's brigade got weapons for combatants\\
A: Yes\\
Missing: The acts of bringing the weapon were made in name of the organisation\\

\textbf{Problem 148}\\
P: Carmine Rocco, Health Director was contacted in June by Sue Robbins, Division of Waste Management, to advise him that she would be in the county to conduct a site visit to the former Woodson-Tenent Laboratory\\
H: Sue Robbins visits the former Woodson-Tenent Laboratory\\
A: Yes\\
Missing: She carried out her intention\\

\textbf{Problem 149}\\
P: In 1869 Sumner resigned his see, but continued to live at the official residence at Farnham until his death on the 15th of August 1874\\
H: Sumner died at Farnham\\
A: No\\
Reason: he might have died somewhere other than his residence\\

\textbf{Problem 150}\\
P: Paralysis was followed by aphasia, and after acute pain, followed by a long period of apathy, from which death relieved Swift in October 1745\\
H: Swift died in 1745\\
A: Yes\\

\textbf{Problem 154}\\
P: New research shows there has been a sharp increase in disfiguring skin cancers, particularly in women under the age of 40, providing more evidence that young people are not heeding warnings about the dangers of tanning\\
H: Tanning may cause skin cancers\\
A: Yes\\

\textbf{Problem 155}\\
P: Nokia, Texas Instruments and other leading makers of mobile phones have formally complained to Brussels that Qualcomm, the US mobile chipmaker, has unfairly used its patents on 3G technologies\\
H: Texas Instruments produces mobile phones\\
A: Yes\\

\textbf{Problem 166}\\
P: In Italy, big protests by students and university staff against government reforms to higher education brought parts of central Rome to a standstill on Tuesday\\
H: The Italian government introduces reforms to higher education\\
A: No\\
Reason: There is no evidence in the premises that the reforms were actually carried out\\
Reason: because the protests might be against reforms that have merely been suggested\\

\textbf{Problem 167}\\
P: The bus, which was heading for Nairobi in Kenya , crashed in the Kabale district of Uganda near the Rwandan border\\
H: The Kabale district borders on Rwanda\\
A: Yes\\
Lexicon: if “to be near border” can be understood as “to border”\\

\textbf{Problem 172}\\
P: Traditionally, the Brahui of the Raisani tribe are in charge of the law and order situation through the Pass area. This tribe is still living in present day Balochistan in Pakistan\\
H: The Raisani tribe resides in Pakistan\\
A: Yes\\
Lexicon: if X live in Y then X reside in Y\\
Note: this is not obvious however since P is compatible with only (possibly minor) parts of the tribe living in Baluchistan, whereas H seems to require the whole, or at least the vast majority, of the tribe living there\\

\textbf{Problem 173}\\
P: The Armed Forces Press Committee (COPREFA) admitted that the government troops sustained 11 casualties in these clashes, adding that they inflicted three casualties on the rebels\\
H: Three rebels were killed by government troops\\
A: Yes\\
Lexicon: 'casualty' imply death in this context\\

\textbf{Problem 174}\\
P: Four US cable companies, including industry leaders Comcast Corp and Time Warner Cable, have entered the fast-growing wireless arena through a joint venture with Sprint Nextel\\
H: Time Warner Cable is a partner of Sprint Nextel\\
A: Yes\\
Missing: The joint venture is still active\\

\textbf{Problem 176}\\
P: Tom Online revenues for the three months ended June 30, 2005 were US\$42.78 mn, an increase of 38.5\% over the same period in 2004\\
H: Tom Online earned US\$42.78 mn in three months\\
A: No\\
Reason: if the costs are most ofter not zero. (earnings are revenues minus costs)\\

\textbf{Problem 177}\\
P: Tom Online revenues for the three months ended June 30, 2005 were US\$42.78 mn, an increase of 38.5\% over the same period in 2004\\
H: Tom Online revenues have risen by 38.5\%\\
A: No\\
Reason: tense mismatch\\

\textbf{Problem 178}\\
P: Anglo/Dutch Royal Dutch Shell, Total, of France, and Spain's Repsol were all named as examples of established oil companies involved in the oil-for-food programme before surcharges began in 2001\\
H: Total participated in the oil-for-food programme\\
A: Yes\\
Claim: The naming is truthful\\

\textbf{Problem 187}\\
P: GUS on Friday disposed of its remaining home shopping business and last non-UK retail operation with the €390m (£265m) sale of the Dutch home shopping company, Wehkamp, to Industri Kapital, a private equity firm\\
H: Wehkamp cost €390m\\
A: Yes\\

\textbf{Problem 191}\\
P: Though Wilkins and his family settled quickly in Italy, it wasn't a successful era for Milan, and Wilkins was allowed to leave in 1987 to join French outfit Paris Saint-Germain\\
H: Wilkins departed Milan in 1987\\
A: No\\
Reason: Wilkins was only allowed to leave\\

\textbf{Problem 195}\\
P: Zapatero visited the following cities in four days: Brasilia, São Paulo, Buenos Aires and Santiago de Chile. According to official sources these visits are the last part of the project he began at the EU-Latin American Summit in Guadalajara, Mexico and pursued in the Ibero-American meeting in Costa Rica in November\\
H: Zapatero participated in the Ibero-American meeting in Costa Rica\\
A: Yes\\

\textbf{Problem 202}\\
P: "Glue sniffing" is most common among teenagers. They generally grow out of it once other drugs such as alcohol and cannabis become available to them. Seven-year-olds have been known to start "glue sniffing". Because of the social stigma attached to "glue sniffing" most snifters stop around 16 or 17 years, unless they are seriously addicted\\
H: Glue-sniffing is common among youngsters\\
A: No\\
Reason: it might still be incredibly rare in all age groups\\

\textbf{Problem 212}\\
P: Actor Christopher Reeve, best known for his role as Superman, is paralyzed and cannot breathe without the help of a respirator after breaking his neck in a riding accident in Culpeper, Va., on Saturday\\
H: Christopher Reeve had an accident\\
A: Yes\\

\textbf{Problem 221}\\
P: Asia has the highest number of child workers, but Sub-Saharan Africa has the highest proportion of working children relative to population\\
H: Child labor is widely used in Asia\\
A: No\\
Reason: highest proportion does not imply high, and thus, “widely used”, cannot be inferred\\

\textbf{Problem 225}\\
P: Between March and June, scientific observers say, up to 300,000 seals are killed. In Canada, seal-hunting means jobs, but opponents say it is vicious and endangers the species, also threatened by global warming\\
H: Hunting endangers seal species\\
A: Yes\\
Missing: The opponents of seal-hunting are right\\

\textbf{Problem 230}\\
P: Castro's successful visits to Harlem and the Bronx, his speech to the United Nations General Assembly, and the October 21 march of 3,000 against U.S. policy and subsequent picket lines at Cuba's UN mission in defense of the revolution - all these events dealt blows to the U.S. government's unceasing efforts to isolate and slander Cuba\\
H: Castro visits the UN\\
A: No\\
Reason: because one visit does not justify the habitual aspect implied by the present tense\\

\textbf{Problem 231}\\
P: Catastrophic floods in Europe endanger lives and cause human tragedy as well as heavy economic losses\\
H: Flooding in Europe causes major economic losses\\
A: No\\
Reason: Not all floods are catastrophic\\
Missing: Heavy economic losses are major ones\\

\textbf{Problem 232}\\
P: China's announcement of a rival Panchen Lama to the boy already recognized by the Dalai Lama are indications that its Communist regime is desperate in using any means possible to strengthen control over Tibet and Tibetan affairs\\
H: Dalai Lama and the government of the People's Republic of China are in dispute over Panchen Lama's reincarnation\\
A: Yes\\
World: The successive Panchen Lamas form a tulku reincarnation lineage which are said to be the incarnations of Amitābha\\
World: There can only be one Panchen Lama reincarnation at a time according to either the Dalai Lama or the Chinese Government\\

\textbf{Problem 278}\\
P: In November 1990, the president announced that opposition political parties would be permitted to organize in 1991. Several new parties emerged, including the Democratic Republican Movement (MDR), the Liberal Party (LP), the Democratic and Socialist Party (PSD), and the Coalition for the Defense of the Republic (CDR)\\
H: Several new political parties emerged\\
A: Yes\\
Missing: All parties mentioned are political parties\\
Language: H is formulated in a weird way\\

\textbf{Problem 286}\\
P: Italian film-maker, Fellini was awarded an honorary Oscar for lifetime achievement. He died on October 31, 1993\\
H: An Italian director is awarded an honorary Oscar\\
A: Yes\\
World: A film-maker is a director\\

\textbf{Problem 289}\\
P: Kieslowski later said that he abandoned documentary filmmaking due to two experiences: the censorship of Workers '71, which caused him to doubt whether truth could be told literally under an authoritarian regime, and an incident during the filming of Station (1981) in which some of his footage was nearly used as evidence in a criminal case. He decided that fiction not only allowed more artistic freedom, but could portray everyday life more truthfully\\
H: Kieslowski is a director\\
A: Yes\\
Missing: He has not abandoned filmmaking\\
World: all makers of non-documentary films are directors\\

\textbf{Problem 293}\\
P: Medium-size black holes actually do exist, according to the latest findings from NASA's Hubble Space Telescope, but scientists had to look in some unexpected places to find them. The previously undiscovered black holes provide an important link that sheds light on the way in which black holes grow\\
H: Hubble discovers black holes\\
A: Yes\\
Language: if Telescope findings are classified as discoveries by the telescope itself\\

\textbf{Problem 294}\\
P: Mental health problems in children and adolescents are on the rise, the British Medical Association has warned, and services are ill-equipped to cope\\
H: Mental health problems increase in the young\\
A: Yes\\
Claim: The British Medical Association was telling the truth when they made their warning\\
Lexicon: Children and adolescents are called young\\

\textbf{Problem 295}\\
P: Most people are familiar with the idea of St. Bernards or other dogs taking part in rescue and recovery efforts. Robots might also take part in search and rescue missions\\
H: Robots are used to find missing victims\\
A: No\\

\textbf{Problem 308}\\
P: ON 29 June the Dutch right-wing coalition government collapsed. It was made up of the Christian-democrats (CDA) led by Prime Minister Jan Peter Balkenende, the right wing liberal party (VVD) and the so-called 'left-liberal' D66\\
H: Three parties form a Dutch coalition government\\
A: No\\
Reason: The coalition has collapsed\\

\textbf{Problem 310}\\
P: On July 12 Portuguese President Jorge Sampaio asks Pedro Santana Lopes to form a government. The new government is sworn in on July 17 and includes António Monteiro as foreign minister, Daniel Sanches as interior minister, and António Bago Flix as finance minister; Paulo Portas remains defense minister\\
H: New Portuguese prime minister is elected\\
A: No\\
Reason: (H carries universal meaning)\\

\textbf{Problem 313}\\
P: Over a course of days, the bank went from apparent strength to bankruptcy. Barings was Britain's oldest merchant bank. It had financed the Napoleonic wars, the Louisiana purchase, and the Erie Canal. Barings was the Queen's bank\\
H: Barings was Britain's oldest merchant bank\\
A: Yes\\

\textbf{Problem 316}\\
P: President Chirac has been advised that his resumption of nuclear testing would precipitate an international boycott of French products\\
H: International pressure is exerted to end French nuclear tests\\
A: No\\
Reason: because tests have not been resumed\\
Reason: because “has been advised” is not necessarily a disguise term for international pressure\\

\textbf{Problem 346}\\
P: The first dinosaur remains in south-east Asia were found in Laos as early as 1936, but little else eventuated until the 1980s. Much of the area is heavily vegetated, and although difficult to explore, promises many significant finds in the future. Thailand has provided the best finds, particularly from the Khorat Group, which covers the period from the late Triassic to the early Cretaceous\\
H: Dinosaur remains were found in Asia\\
A: Yes\\

\textbf{Problem 352}\\
P: The increased amounts of carbon dioxide (CO2) and other greenhouse gases (GHGs) are the primary causes of the human-induced component of global warming\\
H: Greenhouse effect changes global climate\\
A: Yes\\
World: Global warming is a change in global climate\\
World: Increased amounts of CO2 and other greenhouse gases cause Greenhouse effect\\

\textbf{Problem 376}\\
P: The son of Italy's last king returned to his homeland Monday, ending more than a half-century of exile. Victor Emmanuel and his family landed at a military airport in Rome in a private plane and headed immediately for an audience with Pope John Paul II\\
H: Italian royal family returns home\\
A: No; (H carries universal meaning)\\

\textbf{Problem 378}\\
P: The U.S. space shuttle Atlantis, its lights flashing like a beacon, smoothly docked with Russia's space station Mir Wednesday on its mission to pick up U.S. astronaut Shannon Lucid, who has spent a record-breaking six months in orbit\\
H: US shuttle Atlantis docks with the Mir space station\\
A: Yes\\
Language: "docks" is understood in the possible mood as opposed to the habitual aspect (this is the most available interpretation to me)\\

\textbf{Problem 396}\\
P: While the Baltic countries are set to join the European Union soon, their economies are already in good shape. Estonia, Latvia and Lithuania boast the fastest economic growth in the entire Baltic Sea region, for example\\
H: The Baltic Countries will join the EU\\
A: No\\
Reason: since one may be set to do X and nevertheless not end up doing X\\

\textbf{Problem 406}\\
P: The Cyrillic alphabet is an alphabet used for several East and South Slavic languages; (Belarusian, Bulgarian, Macedonian, Russian, Rusyn, Serbian, and Ukrainian) and many other languages of the former Soviet Union, Asia and Eastern Europe. It has also been used for other languages in the past. Not all letters in the Cyrillic alphabet are used in every language which is written with it\\
H: Cyrillic is an alphabet used for certain Slavic languages, such as Russian\\
A: Yes\\

\textbf{Problem 408}\\
P: Her Majesty Queen Elizabeth II was born in London on April, 21 1926, first child of the Duke and Duchess of York, subsequently King George VI and Queen Elizabeth. Five weeks later she was christened in the chapel of Buckingham Palace and was given the names Elizabeth Alexandra Mary Windsor. The Queen ascended the throne on February 6, 1952 upon the death of her father, King George VI. Her Coronation followed on June 2, 1953\\
H: Elizabeth the Second's father was George VI\\
A: Yes\\
Lexicon: King George VI can be referred with “George VI”\\

\textbf{Problem 420}\\
P: Yachtsman Melvyn Percy became so fed up with the standard of service when preparing his boat for the 3000-mile Atlantic crossing that he decided to set up his own company on the Clyde to provide Scots sailors with professional advice and practical assistance. Minerva Rigging, based at Kip Marina just south of Gourock, now employs five full-time staff and is the Scottish agent for Kemp, one of Europe's leading yacht spar and mast manufacturers\\
H: Melvyn Percy set up Minerva Rigging at Kip Marina\\
A: Yes\\
Missing: He acted on his decision\\

\textbf{Problem 421}\\
P: Of all the national park lands in the United States, none is closer to a major urban area or more beset with problems than the Everglades, a shallow, 50-mile-wide river of grass that flows south from Lake Okeechobee to Florida Bay. For more than a generation, this fragile natural wonder has been held hostage by a web of special interests -- farmers, sportsmen and about 6 million South Florida residents in need of drinking water and flood control -- that is as complex as the ecosystem itself\\
H: The Everglades is 50-mile wide\\
A: Yes\\

\textbf{Problem 422}\\
P: Aeschylus is often called the father of Greek tragedy; he wrote the earliest complete plays which survive from ancient Greece. He is known to have written more than 90 plays, though only seven survive. The most famous of these are the trilogy known as Orestia. Also well-known are The Persians and Prometheus Bound\\
H: "The Persians" was written by Aeschylus\\
A: Yes\\
Nitpick: We assume coherence is observed\\

\textbf{Problem 434}\\
P: November 9, 1989 , the day the Berlin Wall fell and the world changed forever . Not even the most astute saw it coming . As Hungary's foreign minister in the late summer of 1989 , Gyula Horn gave the order to let visiting East Germans use his country to do a 400-mile end run around the Berlin Wall , a move now seen as the beginning of the end for hard-line communism in Europe\\
H: The Berlin Wall was torn down in 1989\\
A: Yes\\
Lexicon: A falling border and a torn border colloquially refer to the same thing\\

\textbf{Problem 438}\\
P: It would force countries in the region to choose between the United States and Japan -- or maybe between Japan and China . And it would collapse whatever promise the newly born Asia Pacific Economic Cooperation (APEC) holds as a vehicle to assure American commercial access to the booming Pacific markets . To risk all of this in the name of a flawed concept is foolish . This is not to deny the Administration 's understandable frustration with a persistently huge Japanese trade surplus\\
H: APEC is the newly born Asia Pacific Economic Cooperation\\
A: Yes\\

\textbf{Problem 440}\\
P: Article 19 said a correspondent for the independent newspaper Narodnaya Volya who attempted to determine how many people involved in emergency work in the contaminated zone around the Chernobyl nuclear reactor after the 1986 catastrophe were still alive was told by the Ministry of Emergencies that non-state newspapers could be refused access to any information\\
H: The catastrophe at Chernobyl happened in 1986\\
A: Yes\\
O: “the Chernobyl nuclear reactor” entails that was in Chernobyl\\

\textbf{Problem 441}\\
P: After leaving Time-Life, Shuker joined the Public Broadcast Laboratory, forerunner of PBS, and participated in two pioneering films -- 'The Chair,' about an attorney trying to save a condemned man, and 'Free At Last,' a 90-minute documentary about Dr. Martin Luther King Jr. that by chance was being made when King was murdered in 1968\\
H: Martin Luther King was murdered in 1968\\
A: Yes\\

\textbf{Problem 446}\\
P: That theory, a concept first proposed by Albert Einstein and often reiterated by British astrophysicist Stephen Hawking, seeks to find a single set of equations that can explain all the fundamental forces in the universe: gravity, electromagnetism, and the strong and weak interactive forces among subnuclear particles\\
H: Stephen Hawking is a physicist\\
A: Yes\\
Lexicon: astrophysics is a kind of physics\\

\textbf{Problem 451}\\
P: Jurassic Park is a novel written by Michael Crichton. Jurassic Park was published in 1990\\
H: Michael Crichton is the author of the book Jurassic Park\\
A: Yes\\
World: Novels are published as a books\\

\textbf{Problem 452}\\
P: Boris Becker is a true legend in the sport of tennis. Aged just seventeen, he won Wimbledon for the first time and went on to become the most prolific tennis player\\
H: Boris Becker is a Wimbledon champion\\
A: Yes\\
World: The winner of Wimbledon is its champion\\

\textbf{Problem 454}\\
P: On Aug. 6, 1945, an atomic bomb was exploded on Hiroshima with an estimated equivalent explosive force of 12,500 tons of TNT, followed three days later by a second, more powerful, bomb on Nagasaki\\
H: In 1945, an atomic bomb was dropped on Hiroshima\\
A: No\\
Reason: An atomic bomb can explode without being dropped\\

\textbf{Problem 463}\\
P: Qin Shi Huang, personal name Zheng, was king of the Chinese State of Qin from 247 BCE to 221 BCE, and then the first emperor of a unified China from 221 BCE to 210 BCE, ruling under the name First Emperor\\
H: Qin Shi Huang was the first China Emperor\\
A: No\\
Reason: there could be emperors of non-unified china\\

\textbf{Problem 469}\\
P: George Herbert Walker Bush (born June 12, 1924) is the former 41st President of the United States of America. Almost immediately upon his return from the war in December 1944, George Bush married Barbara Pierce\\
H: The name of George H.W. Bush's wife is Barbara\\
A: Yes\\
Missing: George H.W. Bush did not divorce\\

\textbf{Problem 481}\\
P: The Gurkhas come from mountainous Nepal and are extremely tenacious warriors-as foot soldiers they are the best only in close combat jungle/mountain combat and hand to hand type situation\\
H: The Gurkhas come from Nepal\\
A: Yes\\

\textbf{Problem 482}\\
P: Arromanches-les-Bains or simply Arromanches is a town in Normandy, France, located on the coast in the heart of the area where the Normandy landings took place on D-Day, on June 6, 1944\\
H: The Normandy landings took place in June 1944\\
A: Yes\\

\textbf{Problem 484}\\
P: Kohl participated in the late stage of WWII as a teenage soldier. He joined the Christian-Democratic Union (CDU) in 1947\\
H: The name of Helmut Kohl's political party is the Christian Democratic Union\\
A: Yes\\
World: He was elected Chancellor when he was a member of the CDU (as its candidate for that office). The CDU is a political party\\

\textbf{Problem 506}\\
P: The planet probably got this name due to its red color; Mars is sometimes referred to as 'the Red Planet'\\
H: Mars is called 'the red planet'\\
A: No\\
Reason: “sometimes” does not allow such an inference\\

\textbf{Problem 513}\\
P: The Arabic Language is the door to Islam and the Language of the holy Quran and Prophetic traditions. Therefore, distancing ourselves from Arabic studies leads to innovation and disbelief\\
H: Arabic is the language of the Quran\\
A: Yes\\

\textbf{Problem 516}\\
P: The biggest newspaper in Norway, Verdens Gang, prints a letter to the editor written by Joe Harrington and myself\\
H: Verdens Gang is a Norwegian newspaper\\
A: Yes\\
Missing: Verdens Gang is Norwegian, not a foreign paper that happens to be the biggest newspaper in Norway\\

\textbf{Problem 517}\\
P: A number of undergraduate schools and colleges have also received it, as has the French Red-Cross, the abbey of Notre-Dame des Dombes, and the French railway company SNCF\\
H: The French railway company is called SNCF\\
A: Yes\\

\textbf{Problem 536}\\
P: The beleaguered Euro-Disney theme park outside Paris is doing so poorly it might have to close unless it gets help soon from its lenders, the chairman of Walt Disney Co. said in an interview published Friday\\
H: Euro-Disney is a theme park outside Paris\\
A: Yes\\
Nitpick: There is only one Euro-Disney\\

\textbf{Problem 540}\\
P: New Delhi: More than 100 Nobel prize winners, two US congressmen, and leading labour organizations have expressed concern over threats against the life of Kailash Satyarthi, India's leading opponent of child labour\\
H: Kailash Satyarthi, India's leading opponent of child labour\\
A: Yes\\
Remark: Badly formed problem (Interpreting the hypothesis to say Satyarthi *is* the leading opponent)\\

\textbf{Problem 541}\\
P: The hurdler suspended in 1993 for taking the anabolic steroid Nandrolone was cleared after a three judge arbitration panel ruled that the case against her could not be proved beyond reasonable doubt\\
H: Nandrolone is a steroid\\
A: Yes\\

\textbf{Problem 542}\\
P: Even while accepting the Russian plan, IMF Managing Director Michel Camdessus noted that the efficiency of Russia's State Taxation Service "is declining rapidly."\\
H: Michel Camdessus is managing director of IMF\\
A: Yes\\

\textbf{Problem 545}\\
P: Rolf Ekeus the Swedish diplomat who chairs the commission said that while Baghdad had been cooperative in helping his panel establish procedures for monitoring and verifying the weapons destruction it is unclear how the Iraqis will react once the system is in operation\\
H: Rolf Ekeus is a Swedish diplomat\\
A: Yes\\

\textbf{Problem 547}\\
P: For the case of Aldrich Hazen (Rick) Ames , and his Colombian-born wife , Maria del Rosario Casas Ames arrested by the FBI on charges that they were Russian spies , is so rich in mysteries , puzzles , riddles and contradictions that the answers may never be untangled even as the CIA's counterintell\\
H: Aldrich Hazen Ames is married to Rosario Casas Ames\\
A: Yes\\

\textbf{Problem 549}\\
P: Since independence and the blood bath of partition , in which an estimated 15 million people were uprooted and 200,000 died , India and Pakistan have fought three wars , two over possession of Kashmir\\
H: India and Pakistan have fought three wars for the possession of Kashmir\\
A: No\\
Reason: (2 < 3)\\

\textbf{Problem 551}\\
P: New Zealand film director Peter Jackson is to make a 264-million-New Zealand-dollar (133 million U.S. dollars) trilogy based on Tolkien's fantasy classic "Lord of The Rings," the local TV station reported Tuesday evening\\
H: Tolkien wrote the fantasy epic called "Lord of the Rings"\\
A: Yes\\
Language: Using genitive 's' is accepted as a way to ascribe authorship\\
World: "Lord of the Rings" is an epic. The works that films are based on are said to be written\\

\textbf{Problem 553}\\
P: Colin L. Powell and Laura Bush, wife of Gov. George W. Bush, are to speak on the opening night in Philadelphia, while the Democrats have tentative plans to have President Clinton and Hillary Rodham Clinton address the delegates on the first night in Los Angeles\\
H: The name of George W. Bush's wife is Laura\\
A: Yes\\

\textbf{Problem 558}\\
P: In the early hours of April 15, 1912, the British luxury liner Titanic sank in the North Atlantic off Newfoundland, less than three hours after striking an iceberg\\
H: The Titanic sank in 1912\\
A: Yes\\

\textbf{Problem 562}\\
P: In "Killing the Dream: James Earl Ray and the Assassination of Martin Luther King, JR.", by Gerald Posner (Harvest/Harcourt Brace, \$15), the writer who attacked Kennedy conspiracy theories in "Case Closed" takes on Ray's claim of innocence in the 1968 murder of King\\
H: Martin Luther King was murdered in 1968\\
A: Yes\\
Nitpick: “King” is “Martin Luther King”\\

\textbf{Problem 573}\\
P: A seven-member Tibetan mountaineering team conquered the 8,586-meter Mt. Kanchenjunga, the third highest peak of the world, the Chinese mountaineering association confirmed here on Sunday\\
H: Kanchenjunga is 8586 meters high\\
A: Yes\\
Language: An n-meter mountain is is n-meters high\\

\textbf{Problem 579}\\
P: Salma Hayek drew a crowd in Veracruz, Mexico, at the July 8 premiere of 'Nobody Writes to the Colonel', a movie based on a short novel by Nobel laureate Gabriel Garcia Marquez\\
H: Gabriel Garcia Marquez is a Nobel prize winner\\
A: Yes\\

\textbf{Problem 581}\\
P: Two prominent scientists have made guest appearances on the show, paleontologist Stephen Jay Gould and Stephen Hawking, the theoretical physicist whose brilliance has been compared with Einstein's\\
H: Stephen Hawking is a physicist\\
A: Yes\\

\textbf{Problem 586}\\
P: The stupendous power of the Tevatron made possible the 1995 discovery of the top quark - the last of six flavors of quarks predicted by the standard model theory of particle physics\\
H: The top quark is the last of six flavors of quarks predicted by the standard model theory of particle physics\\
A: Yes\\

\textbf{Problem 588}\\
P: Within trillionths of a second after the Big Bang, they have reasoned, the primordial explosion - many billions of degrees hot - must have created a universe that contained equal quantities of matter and antimatter\\
H: The Big Bang is the primordial explosion from which the universe was created\\
A: No\\
Reason: "they reasoned" cannot be taken for fact\\

\textbf{Problem 594}\\
P: Egypt, Syria and the members of the Gulf Cooperation Council, namely Bahrain, Kuwait, Oman, Qatar, Saudi Arabia and the United Arab Emirates, signed the Damascus Declaration after the 1991 Gulf War to create a mechanism aimed at guaranteeing security in the Gulf region\\
H: The members of the Gulf Cooperation Council are: Saudi Arabia, Kuwait, United Arab Emirates, Qatar, Oman, Bahrain\\
A: Yes\\

\textbf{Problem 599}\\
P: Foreign Ministers of the 16 NATO member countries met early this morning with their Russian counterpart, Yevgeny Primakov, in Berlin in a bid to improve NATO-Russia relations\\
H: The NATO has 16 members\\
A: Yes\\

\textbf{Problem 604}\\
P: Mr David Herman, head of GM's local operations accused Mr Gerhardt Schroeder, prime Minister of Lower Saxony and a member of the VW supervisory board, of trying to use his political weight to influence the investigations by state prosecutors in Hesse into claims of industrial espionage against GM\\
H: Gerhardt Schroeder was accused of helping VW\\
A: Yes\\
Missing: The industrial espionage was claimed to have been done by VW\\
Missing: Harming GM benefits VW\\

\textbf{Problem 607}\\
P: The west has preferred to focus on endangered animals, rather than endangered humans. African elephants are hunted down and stripped of tusks and hidden by poachers. Their numbers in Africa slumped from 1.2m to 600,000 in a decade until CITES - the Convention on International Trade in Endangered Species - banned the trade in ivory\\
H: An international convention banned trade in ivory\\
A: Yes\\
Nitpick: CITES is an international convention\\

\textbf{Problem 609}\\
P: Sheriff's officials said a robot could be put to use in Ventura County, where the bomb squad has responded to more than 40 calls this year\\
H: Police use robots for bomb-handling\\
A: No\\
Reason: The Sheriff’s officials only said that they could be used for this purpose. This does not imply habitual use\\

\textbf{Problem 610}\\
P: The researchers in the latest study fed one group of mice a diet in which 60 percent of calories came from fat. The diet started when the mice, all males, were 1 year old, which is middle-age in mouse longevity. As expected, the mice soon developed signs of impending diabetes, with grossly enlarged livers, and started to die much sooner than mice fed a standard diet\\
H: At the age of one year, male mice were fed with a diet in which 60 percent of calories came from fat\\
A: Yes\\

\textbf{Problem 615}\\
P: For example, if you want to find out what the weather in your area will be like for the next five days, select the weather channel from the main multiscreen and you will find detailed forecasts and other weather news. Or if your team has been in action, you can read the top headlines for your chosen sport by keying in 300 on your handset - just as you would on Ceefax\\
H: The weather channel gives you information about the weather in the next few days\\
A: No\\
Reason: Turning on the weather channel implies getting information --- but nothing says that it comes directly from it\\

\textbf{Problem 618}\\
P: Some passengers were escorted off the ship in wheelchairs by crew wearing blue gloves\\
H: Wheelchairs were used to lead passengers off the ship\\
A: No\\
Reasion: lack of determiner is read as universal in this case\\

\textbf{Problem 629}\\
P: Bangladesh, one of the world's poorest countries, suffered a devastating drought and famine in 1974, which killed 1.5 million people. While trying to help starving villagers, he met a 21-year-old woman named Sufia Begum, who was burdened by a tiny yet crushing debt, Yunus recalled in his autobiography, "Banker to the Poor."\\
H: 1.5 million people were killed during the drought and famine devastation in 1974 in Bangladesh\\
A: Yes\\

\textbf{Problem 635}\\
P: Norwegian police released one of three people arrested when Edvard Munch's painting 'The Scream', stolen from the National Gallery in Oslo, was found\\
H: Norwegian police arrested three people for the theft of Munch's painting, 'The Scream'\\
A: No\\
Reason: it is not clear from the premise that the three people were arrested for the theft\\

\textbf{Problem 637}\\
P: In France, art dealers are obliged by law to register all purchases except those bought at a public auction\\
H: In France, works of art bought at public auction are exempt from registration\\
A: Yes\\
World: No one other than art dealers register art works sold at a public auction\\

\textbf{Problem 643}\\
P: The Federal Bureau of Investigation started an independent probe of the circumstances shortly after the White House made plain that President Bill Clinton considered industrial espionage a particular threat to US economic interests\\
H: President Clinton thinks that industrial espionage is a threat to America's well being\\
A: No\\
Reason: The White House is not necessarily being forthcoming about the president's actual beliefs\\

\textbf{Problem 653}\\
P: Argentina's campaign to re-establish itself as a trustworthy trading partner was boosted yesterday with the underwriting of a Dollars 9.5m buyer credit by Britain's Export Credit Guarantee Department\\
H: Britain's Export Credit Guarantee Department underwrites loans to Argentina\\
A: Yes\\

\textbf{Problem 657}\\
P: Argentina announced that it has decided to lift financial and trade restrictions on imports from Britain that were imposed during the 1982 Falklands conflict\\
H: Argentina lifted restrictions on British imports\\
A: Yes\\
Missing: Announcing such a decision is enough to actually lift the restrictions\\

\textbf{Problem 658}\\
P: UK foreign secretary Douglas Hurd will meet President Carlos Menem in Argentina next week\\
H: Foreign Secretary Douglas Hurd visited Argentina\\
A: No\\
Reason: He only is scheduled to visit Argentina in the future (next week)\\

\textbf{Problem 659}\\
P: Britain agreed to lift by March 31 a 150-mile military protection zone enforced around the islands since Argentina invaded them in 1982\\
H: The military protection zone around Falklands was lifted\\
A: No\\
Reason: we do not know if Britain kept its promise; we don't know if March 31 refers to a past date; there may be other islands than Falklands\\

\textbf{Problem 666}\\
P: Relations between Argentina and Britain were soured again last May when Britain decided to extend territorial waters to 200 miles around South Georgia and the South Sandwich Islands\\
H: Britain angered Argentina\\
A: Yes\\
Lexicon: If souring relations imply anger on behalf the involved parties\\

\textbf{Problem 671}\\
P: Although the domestic markets for cigarettes in America in the 1990s was greatly reduced due to bans on advertising, smoking in public places and health warnings, American tobacco companies were showing a higher profit\\
H: Cigarette sales have declined due to restrictions on advertising\\
A: No\\
Reason: it is not clear if advertising has an effect on sales independent of smoking in public places and health warnings\\

\textbf{Problem 672}\\
P: Philip Morris the US food and tobacco group that makes Marlboro, the world's best-selling cigarette, shrugged off strong anti-smoking sentiment in the US\\
H: Philip Morris owns the Marlboro brand\\
A: No\\
Reason: making the product does not imply owning the brand\\

\textbf{Problem 675}\\
P: Gold mining operations in California and Nevada use cyanide to extract the precious metal\\
H: Cyanide is used in gold mining\\
A: Yes\\

\textbf{Problem 677}\\
P: Known as "heap leach" mining, the method has become popular in the last decade because it enables microscopic bits of gold to be economically extracted from low-grade ore\\
H: The mining industry uses a method known as heap leaching\\
A: Yes\\
Clarification: We're talking about the mining industry (as opposed to, say, amature gold hunters)\\

\textbf{Problem 693}\\
P: More than 6,400 migratory birds and other animals were killed in Nevada by drinking water in the cyanide-laced ponds produced by gold mining operations\\
H: Animals have died by the thousands from drinking at cyanide-laced holding ponds\\
A: Yes\\
World: Ponds produced by gold mining operations are called holding ponds\\

\textbf{Problem 704}\\
P: Libya's case against Britain and the US concerns the dispute over their demand for extradition of Libyans charged with blowing up a Pan Am jet over Lockerbie in 1988\\
H: One case involved the extradition of Libyan suspects in the Pan Am Lockerbie bombing\\
A: Yes\\
Nitpick: if one case refers to one of Libya’s cases against Britain and the US\\

\textbf{Problem 710}\\
P: Conservationists fear that one of Namibia's most precious resources, its abundant wildlife and especially its threatened black rhinoceros, faces a major menace from poaching\\
H: In Africa, rhinos are seriously endangered by poaching\\
A: No\\
Reason: P is only about Nambia, not rhinos in general in Africa\\
Reason: Conservationists may not be right\\

\textbf{Problem 713}\\
P: The chaotic development that is gobbling up the Amazon rain forest could finally be reined in with a new plan developed by officials of Amazon countries and leading scientists from around the world\\
H: The Amazon rainforest suffers from chaotic development\\
A: Yes\\
Lexicon: Being gobbled up by X entails suffering from X\\

\textbf{Problem 722}\\
P: As Fiat shows off its robot-controlled inventory stacks in glossy advertisements and IBM switches off the lights in its automated warehouse in Greenock, it is clear that this Cinderella operation has a new set of high-tech glad rags\\
H: Fiat, in particular, uses robots for inventory management\\
A: No\\
Reason: if advertisments not not always depict genuine operational practices\\

\textbf{Problem 723}\\
P: The motor industry accounts for as much as 40 per cent of the 450,000 installed industrial robots worldwide but their use is changing and applications are expanding\\
H: The most common use for robots is the manufacture of automobiles\\
A: No\\
Reason: if 40\% is not enough to license “most common use”\\

\textbf{Problem 728}\\
P: The Arak plant, along with the discovery of a secret Iranian enrichment program in 2003, Tehran's refusal to cease uranium enrichment and findings by IAEA inspectors have increased suspicions about Iran s program\\
H: Iran's program is under suspicion because of the findings by IAEA inspectors\\
A: Yes\\
Language: the context indicates that "because" can be understood to indicate a contributory cause\\

\textbf{Problem 733}\\
P: Mr Lopez Obrador was "sworn in" by Senator Rosario Ibarra, a human rights activist and member of his party, who placed a red, green and white presidential sash across his shoulders\\
H: Senator Rosario Ibarra is a human rights militant\\
A: Yes\\
Lexicon: You can call an activist a militant\\

\textbf{Problem 740}\\
P: Al-Qaida-linked militants have carried out a series of suicide bombings targeting Western interests in Indonesia since 2002\\
H: Since 2002 Al-Qaida militants have tried to hit Western interests in Indonesia\\
A: No\\
Reason: Not all Al-Qaida-linked militants are Al-Qaida militants\\

\textbf{Problem 744}\\
P: Former Prime Minister Rafik Hariri, also a prominent anti-Syria political figure, was killed in a suicide bombing in February last year, which led to rising anti-Syrian waves and the withdrawal of Syrian troops from Lebanon\\
H: Syrian troops have been withdrawn from Lebanon after the murder of Rafik Hariri\\
A: Yes\\

\textbf{Problem 746}\\
P: The Democrats' success in the 2006 elections means changes at the top in the House and Senate\\
H: Democrats won the 2006 elections\\
A: No\\
Reason: success in an election does not imply winning the election\\

\textbf{Problem 750}\\
P: Preliminary tests identified the source of the outbreak as the highly contagious norovirus, which had struck several guests just before they boarded the cruise Nov. 3 in Rome, Carnival officials said\\
H: Norovirus is the source of the outbreak on the ship\\
A: Yes\\
Claim. Preliminary tests were correct\\

\textbf{Problem 754}\\
P: A team from the U.S. Centers for Disease Control boarded the ship when it docked in St. Maarten to oversee the cleaning operation and try to determine what caused the outbreak, Carnival said\\
H: The causes of the outbreak were searched for by a team from the U.S. CDC\\
A: No\\
Reason: if Carnival's report may not be truthful\\
Reason: we only know that they boarded the ship with these intentions and not whether the latter were actually carried\\

\textbf{Problem 756}\\
P: The contaminated pills included metal fragments ranging in size from "microdots" to portions of wire one-third of an inch long, the FDA said\\
H: The contaminated pills contained metal fragments\\
A: Yes\\
Claim: The FDA is telling the truth\\

\textbf{Problem 757}\\
P: Consumers who take any of the contaminated pills could suffer minor stomach discomfort or possible cuts to the mouth and throat, the FDA said, adding that the risk of serious injury was remote\\
H: Contaminated pills could cause minor stomach discomfort or possible cuts to the mouth and throat\\
A: Yes\\
Claim: The FDA is telling the truth\\

\textbf{Problem 758}\\
P: The drug, along with aspirin and ibuprofen, is one of the most widely used pain relievers available without a doctor's note\\
H: Aspirin is one of the most widely used pain relievers available without a doctor's note\\
A: Yes\\

\textbf{Problem 763}\\
P: PST Chittagong, Bangladesh-- Muhammad Yunus, Bangladesh's "Banker to the Poor" who provides loans to help millions of people fight poverty by starting businesses, has won the Nobel Peace Prize\\
H: Muhammad Yunus won the Nobel Prize for Peace\\
A: Yes\\

\textbf{Problem 767}\\
P: "I've always said he's the closest I will ever come to meeting Gandhi, he's simply unmoved by any obstacle or any argument," said Bill Clapp, an heir to the Weyerhaeuser fortune and founder of Global Partnerships, a Seattle microfinance group that has partnered with Grameen in Central America\\
H: Bill Clapp is the founder of the Global Partnerships\\
A: Yes\\

\textbf{Problem 771}\\
P: In 2003, Yunus brought the microcredit revolution to the streets of Bangladesh to support more than 50,000 beggars, whom the Grameen Bank respectfully calls Struggling Members\\
H: Yunus supported more than 50,000 Struggling Members\\
A: Yes\\
Language: Struggling Members is introduced as a proper noun in P\\

\textbf{Problem 784}\\
P: Three nurses in one of Vienna's oldest hospitals have been arrested on suspicion of killing 35 patients\\
H: Three Vienna nurses are under suspicion for killing patients\\
A: Yes\\

\textbf{Problem 792}\\
P: The Supreme Court said today states may bar the removal of life-sustaining treatment from comatose patients who have not made or cannot make their desires known\\
H: There is a Supreme Court decision about the removal of life-support\\
A: Yes\\

\textbf{Problem 800}\\
P: US Steel could even have a technical advantage over Nucor since one new method of steel making it is considering, thin strip casting, may produce higher quality steel than the Nucor thin slab technique\\
H: US Steel may invest in strip casting\\
A: Yes\\
Missing: US Steel is not already investing in strip casting\\

}
\end{document}